\documentclass[letterpaper, 10 pt, conference]{IEEEtran}
\usepackage{amsfonts}
\usepackage{amsmath}
\usepackage{booktabs}
\usepackage{subcaption}
\usepackage{array}
\usepackage[labelfont=normalfont,textfont=normalfont]{caption}
\usepackage[hyphens]{url}
\usepackage[hidelinks]{hyperref}

\usepackage{etoolbox}

\IEEEoverridecommandlockouts

\usepackage{graphicx}
\usepackage{stfloats}

\usepackage{natbib}

\title{\LARGE \bf
Imitation Learning with Precisely Labeled Human Demonstrations
}

\author{Yilong Song}

\begin{document}

\maketitle
\thispagestyle{empty}
\pagestyle{empty}

\begin{abstract}
Within the imitation learning paradigm, training generalist robots requires large-scale datasets obtainable only through diverse curation. Due to the relative ease to collect, human demonstrations constitute a valuable addition when incorporated appropriately. However, existing methods utilizing human demonstrations face challenges in inferring precise actions, ameliorating embodiment gaps, and fusing with frontier generalist robot training pipelines. In this work, building on prior studies that demonstrate the viability of using hand-held grippers for efficient data collection, we leverage the user’s control over the gripper’s appearance---specifically by assigning it a unique, easily segmentable color---to enable simple and reliable application of the RANSAC and ICP registration method for precise end-effector pose estimation. We show in simulation that precisely labeled human demonstrations on their own allow policies to reach on average $88.1\%$ of the performance of using robot demonstrations, and boost policy performance when combined with robot demonstrations, despite the inherent embodiment gap.

\end{abstract}

\section{Introduction}

Eyeing the success of large language models (LLMs), we are increasingly interested in exploring end-to-end vision-language-action (VLA) models as a path to generalist robots. If the LLM's success can be attributed to two factors: expressive architectures and diverse high quality datasets, the VLA's current obstacle lies in obtaining the latter. Robot demonstrations, the most straightforward and proven form of training data, are expensive to collect; \citet{rt12022arxiv}, for instance, used $17$ months to collect $130K$ trajectories, but still fell short in matching the data quantity LLMs use to achieve generality. Methods for collecting and using alternative categories of data are thus explored out of necessity.

Among these alternative categories, human demonstrations are of special importance due to its direct relevance to robot tasks. One category of prior work that seeks to leverage internet-scale human videos pretrains with a self-supervised objective a model that learns a discrete latent action representation from unstructured videos, then maps latent actions to true actions with small amounts of labeled video \citep{LAPO, LAPA}. This class of methods, which we call latent-action methods, is advantageous in its easy setup (involving no hardware usage), wide applicability (to almost any video), scalability (which is the result of the former two properties), and effectiveness in learning representations of motions \citep{bjorck2025gr00t}, but as the tradeoff struggles with higher dimensional continuous action spaces and fine-grained motion planning, and also suffers from shortages of actual large-scale high-quality training data despite inherent scalability \citep{LAPA}. Another category of work finds a middle ground between teleoperated robot data and latent-action data in terms of scalability, control over video content, and label precision by proposing data collection methods that circumvent robot usage \citep{song2020grasping, young2020visual, shafiullah2023bringing, chi2024universal, lepert2025phantomtrainingrobotsrobots}. Our work falls within this category.

All methods utilizing human demonstrations need to address two issues: the lack of action labels, and the presence of an embodiment gap. Our work relies on the observation that, when granting two assumptions: first the appearance of the end-effector stays constant between training and testing (i.e. the same gripper is used hand-held and mounted on the robot), and second an embodiment invariant wrist camera view is provided, a visuomotor policy model trained on \textit{precisely labeled} human demonstrations generalizes zero-shot to the robot embodiment with respectable performance. On this basis, we 
\begin{enumerate}
    \item devise a method for reliable, precise action labeling, and
    \item conduct experiments in simulation quantifying policy performance when trained both solely on precisely labeled human demonstrations and a mixture of human demonstrations and a small number of robot demonstrations,
\end{enumerate}
which can be considered our two contributions. Specifically, balancing cost of implementation and the requirement for precision, we leverage our control over the gripper's appearance by assigning it a unique, easily segmentable color to facilitate robust application of Random Sample Consensus (RANSAC) and Iterative Closest Point (ICP) registration for precise labeling via pose estimation. We show that policies trained on human demonstrations reach on average $88.1\%$ of the performance of the same architecture trained on the same number of robot demonstrations, while the addition of precisely labeled human demonstrations to a training set of robot demonstrations significantly boosts policy performance.

\section{Related Work}
\begin{figure*}[t]
  \centering
  \begin{minipage}[t]{0.49\textwidth}
    \centering
    \includegraphics[width=\textwidth, trim=0 150 0 150, clip]{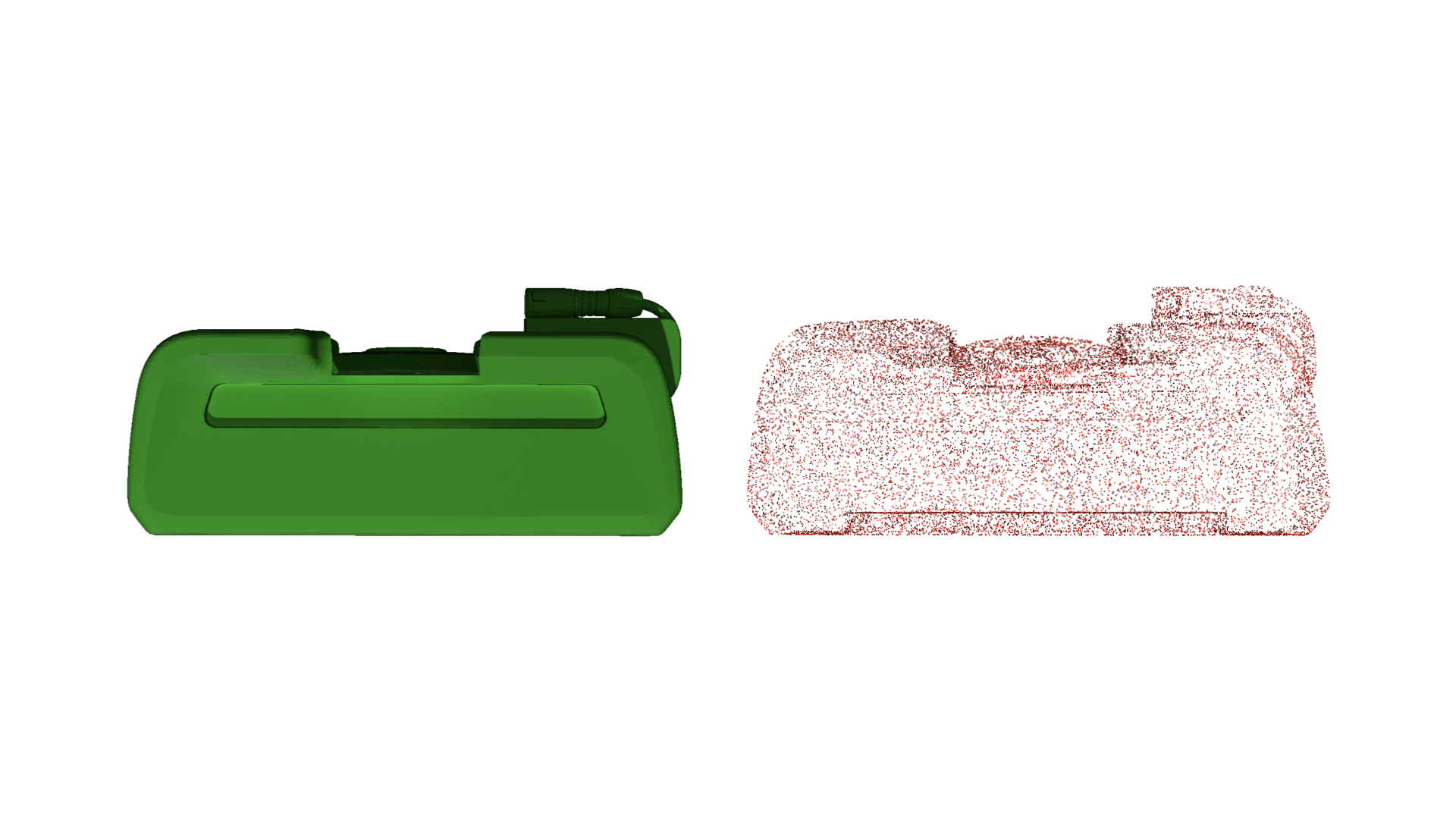}
  \end{minipage}\hfill
  \begin{minipage}[t]{0.49\textwidth}
    \centering
    \includegraphics[width=\textwidth, trim=0 150 0 150, clip]{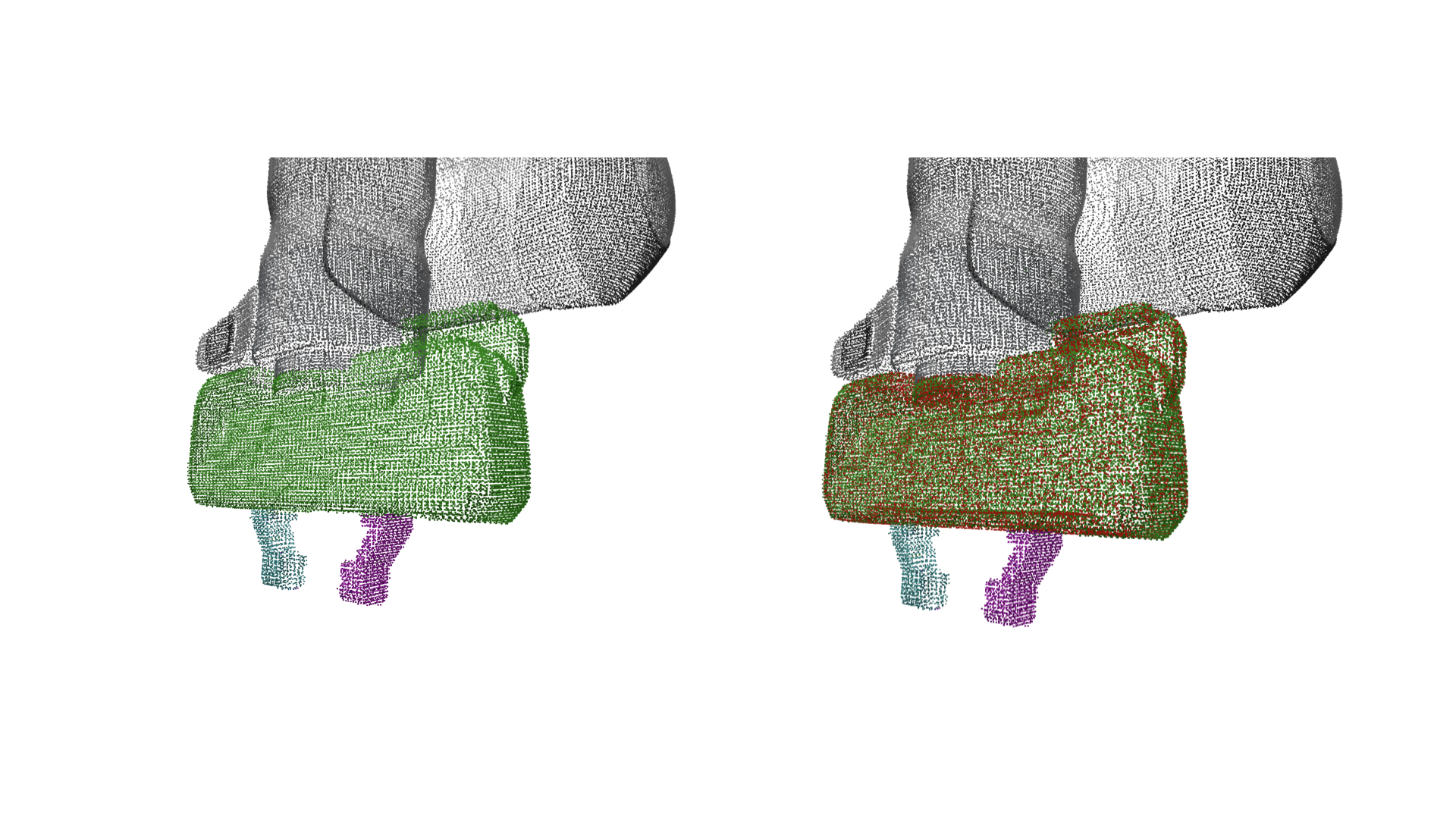}
  \end{minipage}
  \caption{Going from left to right: 1) The triangle mesh of the recolored panda gripper used in simulation; 2) the corresponding point cloud sampled uniformly from the mesh, which we use for pose estimation; 3) a point cloud of the robot (cropped from the scene for clarity) in a simulated demonstration; 4) Random Sample Consensus (RANSAC) and Iterative Closest Point (ICP) are applied to estimate a rigid body transformation that aligns the end-effector point cloud obtained from the mesh and the robot point cloud (i.e. the pose), resulting in precise and reliable end-effector pose estimation without the need to train a deep learning model.}
  \label{fig:combined_figure}
\end{figure*}
\subsection{Learning from Human Demonstrations}
As previously mentioned, the purpose of this category of work is to expand the robotics training set, and the two inherent obstacles every work in this category must in some way address are (1) the lack of action labels, and (2) the presence of an embodiment gap. It is interesting to observe how the two issues are addressed by different existing approaches.

One class of approaches, which we call latent action methods, uses a self-supervised objective to pretrain a model that learns a latent action representation from videos \citep{LAPO, LAPA}. This approach addresses problem (1) by supplying motion-encoded pseudo-labels to human videos, and relies on separate mechanisms to establish correspondences between latent actions and real actions \citep{LAPA}, and similar motions on different embodiments \citep{bjorck2025gr00t}, thus does not address problem (2) directly. This approach is both highly scalable and effective in learning embodiment-insensitive motion representations \citep{bjorck2025gr00t}, but as the tradeoff does not provide precise guidance for learning fine-grained motion planning. Although also involving representation learning, self-supervised objectives, and internet-scale videos, vision encoders tailored for robotics are pretrained to capture visual semantic priors and sometimes state representations instead of just action representations \citep{Radosavovic2022, nair2022r3m, ma2022vip}. A related and also in principle highly scalable class of methods is reward-learning methods, which involves training reward functions that measure the progress or success of a trajectory, sometimes with respect to a language instruction \citep{concept2robot, Chen_2021, ma2022vip}. All three categories share the properties of not explicitly providing action labels or aligning embodiments, but rather assist other modules of the system in these respects implicitly. Our work is similar to them in that we also do not explicitly align embodiments---although we do focus on providing fine-grained, precise, true action labels---and can be easily integrated with large scale imitation learning pipelines.

Less scalable methods that use structured, often locally-collected human manipulation videos, on the other hand, tend to attempt addressing both issues simultaneously. 

Flow-based methods learn to predict the flows of points across the frame sequence, then use the predicted flow to guide a downstream policy \citep{LbW, xu2024flow, ATM, ren2025motion, papagiannis2025rplusx}. Since flows both describe movements and abstract away the appearance of agents and objects, they address problem (1) and (2) simultaneously---but not completely, since as long as points on the agent is tracked, a morphology gap remains as humans and robots move in different ways. Compared to flow-based methods, our method contains fewer modules, no off-the-shelf components, can capture 3D motion more precisely, and can integrate with large-scale imitation learning frameworks easier. 

Object-centric methods \citep{hsu2024spot, Heppert_2024, bahety2024screwmimic, zhu2024visionbased} circumvent problem (2) and provide at least post-grasp solution to problem (1) by tracking objects instead of embodiments, but compared to our method are more constrained in terms of the range of tasks and types of objects they can effectively handle, and cannot easily integrate with large-scale imitation learning systems.

Another prominent class of methods, which we call behavior-prior learning methods, learns motion or planning priors, then uses these priors to improve policy performance and generalization \citep{Bahl_2022, wang2023mimicplay, xu2023xskill, Bharadhwaj_2024, jain2024vid2robot}. These methods provide either implicit (latent) guidance \citep{wang2023mimicplay, xu2023xskill, Bharadhwaj_2024, jain2024vid2robot} or sparse labels via waypoints \citep{Bahl_2022}, while our work focuses on providing precise labels at reasonable cost. Methods in this category address the embodiment gap problem in various ways: \citet{wang2023mimicplay} do so by decoupling high-level latent planning (learned from human play data) from low-level control (trained on robot demonstrations); \citet{jain2024vid2robot} by leveraging paired human-robot trajectory data; \citet{xu2023xskill} by learning a shared cross-embodiment skill representation through self-supervised clustering. These methods require either extra training \citep{wang2023mimicplay, xu2023xskill} or extra idiosyncratic data collection \citep{jain2024vid2robot}; in contrast, our work closes the embodiment gap simply with a common end-effector and does not require the end-to-end policy learning pipeline to be modified.

Lastly, video prediction methods train or fine-tune video generators that predict subsequent frames conditioned on the first frame, and use the predicted frame sequence at inference time as temporally dense motion guidance for the robot \citep{UniPi, AVDC, dreamitate}. If the human at training time and robot at test time use end-effectors with shared appearances, which is the key insight leveraged by \citet{dreamitate}, the action trajectory can be precisely extracted via end-effector pose estimation, bypassing the embodiment difference, solving both problem (1) and (2) simultaneously. Video prediction methods have many advantages, but also three flaws. First is the computational cost of video generation models, which hinders closed-loop control. Second is that the video generation models hallucinate. Third is the prohibitively large extra expense required to enable end-effector pose estimation with precision matching that of our method. Our work uses 3D scene reconstruction to maximize pose estimation precision, while 3D video generation or multi-camera 2D video generation are very computationally involved. Our work is inspired by the use of shared end-effectors in \citet{dreamitate}, but involves no expensive video prediction.

\subsection{Human Demonstration Collection Methods}
Human demonstrations are labeled or unlabeled videos of human agents performing robotics tasks without robots. Compared to robot demonstrations, human demonstrations avoid two expensive factors: robot hardware and teleoperation. Therefore, although data still scales linearly with human labor, the cost of collecting human demonstrations is significantly lower than collecting robot demonstrations. \citet{lepert2025phantomtrainingrobotsrobots} propose collecting videos of humans completing manipulation tasks with their hands, track the hands' poses, then overlay virtual renderings of the target embodiment accordingly, thus addressing both the label and the embodiment problems. Our method in contrast uses a hand-held end-effector, which trades off the convenience of using less hardware for simple and more precise pose estimation. Different from \citet{lepert2025phantomtrainingrobotsrobots}, \citet{song2020grasping, young2020visual, shafiullah2023bringing, chi2024universal} proposed the more portable solution of using hand-held hardware with ego-centric camera angles, which is advantageous for in-the-wild data collection. On the basis of prior work, \citet{chi2024universal} expanded the set of operable tasks by ameliorating occlusion with fisheye lenses and side mirrors, and achieved sub-centimeter action precision with a visual-inertial SLAM pipeline incorporating video and IMU data, which slightly increases cost and decreases portability \citep{chi2024universal}. \citet{chi2024universal} constitute a thoroughly reasoned and well-tested real-world data collection solution, while our work, with its smaller scope, loosens the ``no external camera" constraint and focus on devising and testing a minimalist precise labeling strategy in simulation.

\section{Method}
We follow a simple paradigm: given unlabeled human demonstrations, we label the human demonstrations via end-effector pose estimation, and use the labeled human demonstrations---either on their own or in concatenation with robot demonstrations---to train a visuomotor policy. To retain portability and cost-effectiveness, we devise a method that requires only accessible hardware, namely a 3D printed hand-held gripper and one or more RGB-D cameras, while maintaining precision.

\subsection{Point Cloud Reconstruction}
To most accurately capture 3D geometries in the scene, particularly that of the end effector, we deploy external depth cameras with known matrices and perform point-cloud reconstruction to obtain 3D scenes corresponding to each frame. The reconstruction method is standard.

We start with an unlabeled dataset
$\mathcal{D} = \{ \text{demo}_i \}_{i=1}^{N}$
of human demonstrations. With \( T_i \) denoting the number of timesteps, each demonstration $\text{demo}_i = \left( I_{i,1}, I_{i,2}, \dots, I_{i,T_i} \right)$, where observation \( I_{i,t} \in \mathbb{R}^{V \times 4\times H \times W}\) represents \(V\) RGBD images from \(V\) camera angles with height $H$ and width $W$, is an ordered sequence of observations. For each camera \( c \in \{1, \dots, V\} \) in each observation $I_{i,t}$, denoted $I_{c,i,t}$, every pixel with coordinates \((u,v)\) in $I_{c,i,t}$ with depth value \(d_{c,i,t}(u,v)\) corresponds to a homogeneous 3D point $\mathbf{X}_{c,i,t} = [x, y, z, 1]^T$ in the global frame via:
\begin{equation*}
\mathbf{X}_{c,i,t} = \mathbf{T}_{c}\ h\left(d_{c,i,t}(u,v)\mathbf{K}_{c}^{-1}
\begin{bmatrix}
u \\
v \\
1
\end{bmatrix}\right),
\end{equation*}
where $\mathbf{K}_c$ and $\mathbf{T}_c$ are respectively the intrinsic and extrinsic matrix of camera $c$, and $h$ is the homogenization operator (i.e. $h(\mathbf{X})=\left[\mathbf{X}^T, 1\right]^T$). We apply this to each view at each time step to obtain a series of partial point clouds, and merge partial point clouds to obtain fuller point clouds.

\subsection{End-Effector Pose Estimation}
Inspired by \citet{dreamitate, chi2024universal}, we use a custom end-effector that work both hand-held and mounted on a robot. The design process naturally produces CAD models of the used end-effectors, which, along with the aforementioned point clouds, allows us to apply a simple and reliable model-based pose estimation algorithm that involves no training and minimal tuning.

Concretely, given the point clouds and a CAD model of the end-effector, we first segment out points belonging to the end-effector, then align the CAD model of the end-effector onto the points by estimating the best rigid-body transformation, i.e., the pose. To enable easy segmentation, we simply customize the end-effector to have a distinct color, e.g. green, and segmentation can be performed reliably by simply filtering out points with the color, and taking the largest cluster. To estimate the best rigid-body transformation, we first obtain a coarse alignment using the Random Sample Consensus (RANSAC) algorithm, and then refine this estimate with the Iterative Closest Point (ICP) algorithm. Since videos are temporally dense and adjacent frames capture fine motions, this ``RANSAC plus ICP" combination needs only be applied for the first frame, and the estimated pose of the previous frame can be used as the initial coarse alignment for each subsequent frame to both increase efficiency and easily enforce rotational consistency for symmetric end-effectors. See Figure \ref{fig:combined_figure} for a visualization of this process.

This algorithm labels each frame with the corresponding absolute end-effector pose. To obtain an action-labeled dataset, we simply shift the series of poses forward by one time step so that the estimated pose at frame $t+1$ becomes the goal pose at frame $t$.

\begin{table*}[htbp]
    \centering
    \small
    \caption{\textbf{Downstream Policy Performance.} We report the success rates of the visual diffusion policy on different data mixtures, averaged across 50 different environment initial conditions. Each cell contains (best single test success rate)/(average success rate of best consecutive 10 checkpoints)/(starting epoch of the best consecutive 10 checkpoints). The best results in each column are in bold; the second best is underlined.}
    \label{tab:robosuite_results}
    \begin{tabular}{ l | c | c | c | c | c }
        \toprule
         & Square $D_0$ & Coffee $D_1$ & Stack $D_0$ & Threading $D_0$ & Three Piece Assembly $D_0$ \\
        \midrule
        $50\text{TD}+0\text{HD}$  & 0.70/0.63/0650  & 0.20/0.12/0050  & 0.74/0.48/0050  & 0.66/0.57/0200  & 0.38/0.30/0050 \\
        $200\text{TD}+0\text{HD}$  & \textbf{0.98}/\textbf{0.94}/2550  & \underline{0.76}/0.63/0050  & \textbf{1.00}/\textbf{0.98}/0050  & \textbf{0.96}/\underline{0.89}/0650  & \underline{0.78}/\underline{0.71}/1400 \\
        \midrule
        $0\text{TD}+200\text{HD}$  & 0.86/0.82/1250  & 0.70/\underline{0.65}/0100  & 0.76/0.69/2550  & 0.02/0.00/0050  & 0.74/0.65/0150 \\
        \midrule
        $50\text{TD}+50\text{HD}$  & 0.74/0.69/1250  & 0.46/0.26/0050  & 0.72/0.61/0050  & 0.84/0.70/0200  & 0.58/0.48/0050 \\
        $50\text{TD}+100\text{HD}$  & 0.80/0.71/0100  & 0.54/0.41/0100  & 0.84/0.79/0050  & 0.86/0.79/0650  & 0.68/0.59/0600 \\
        $50\text{TD}+200\text{HD}$  & 0.88/0.83/0050  & 0.68/0.63/0050  & \underline{0.96}/\underline{0.91}/0750  & 0.94/0.85/0550  & 0.74/0.68/0550 \\
        $50\text{TD}+400\text{HD}$  & \underline{0.94}/\underline{0.88}/0350  & \textbf{0.88}/\textbf{0.81}/0100  & 0.94/0.88/1350  & \textbf{0.96}/\textbf{0.93}/1000  & \textbf{0.82}/\textbf{0.74}/0150 \\
        \bottomrule
    \end{tabular}
\end{table*}

\section{Experiments}
We evaluate our approach in robosuite simulated environments, with diffusion policy \citep{chi2024diffusionpolicy} as the downstream behavorial cloning architecture of choice. Specifically, we choose MimicGen \citep{mandlekar2023mimicgen} tasks due to the larger datasets MimicGen provides (1000 demonstrations each task that can be downloaded directly and the flexibility to generate more demonstrations across multiple embodiments). MimicGen datasets are each generated from 10 source human teleoperated demonstrations by segmenting object-centric subtasks and leveraging trajectory equivariance in relations to new object poses, thus are different from fully human collected datasets. But since \citet{mandlekar2023mimicgen} demonstrated for the exact tasks experimented in this work that policy performance on MimicGen data ``can be comparable to performance on an equal amount of human demonstrations," and that increasing the number of MimicGen generated demonstrations in the training set has a diminishing return on performance, we conclude that the use of MimicGen datasets does not significantly impact trends observed in our results, at least not to our advantage, and that the usage is reasonable. MimicGen datasets are arranged into categories $D_0, D_1, D_2$ indicating different reset distribution variabilities in the object poses, with $D_0$ being the baseline reset distribution, and $D_1$ and $D_2$ having increasing variability. We choose to use the standard $D_0$ reset distribution for all tasks except Coffee, because for Coffee $D_0$ the diffusion policies reach a test success rate of $1$ across all data mixtures quickly, and does not reveal useful information regarding our objective.

\begin{figure}[t] % The [H] option forces the figure to appear exactly here (requires the float package)
  \centering
  \includegraphics[width=0.48\textwidth]{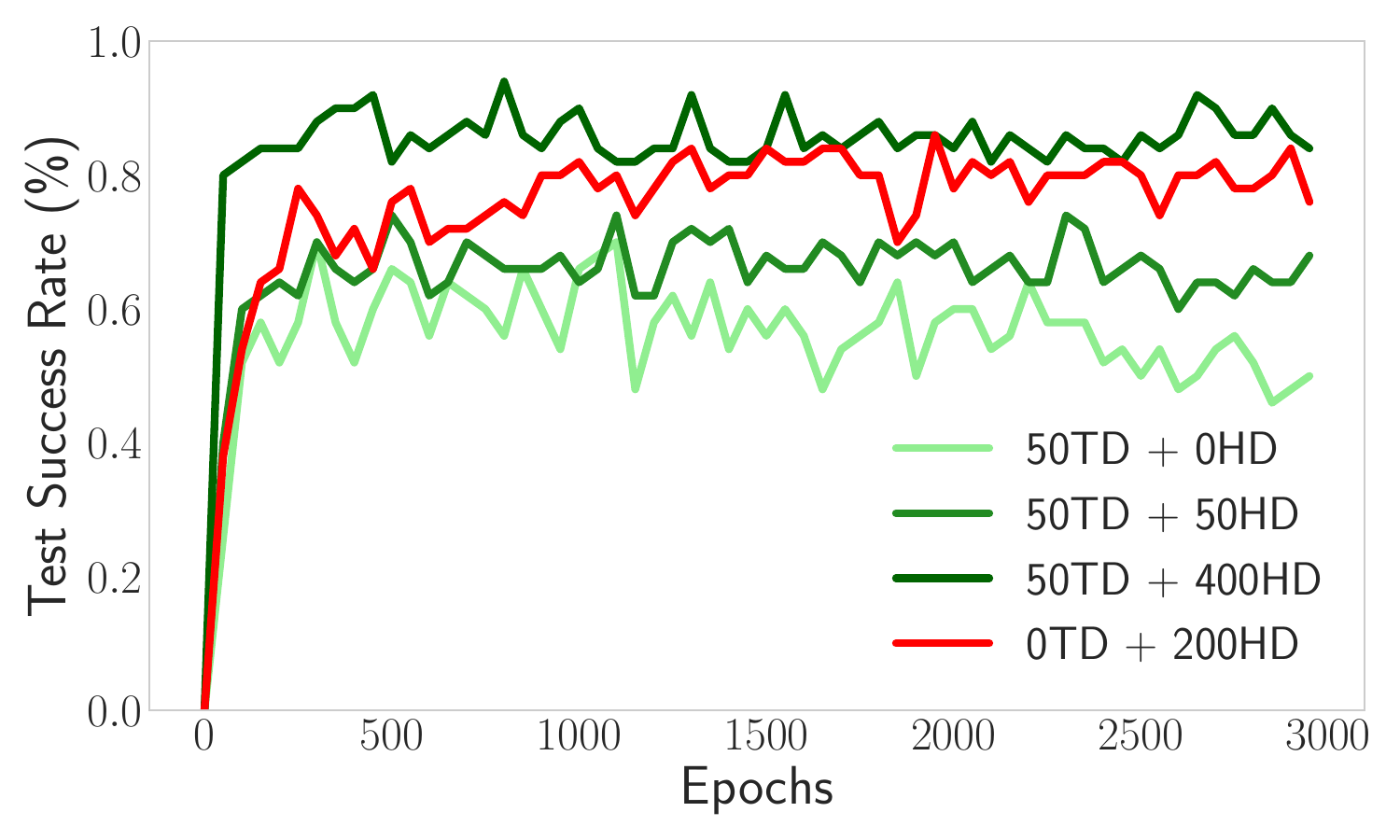}
  \caption{\textbf{Downstream Policy Performance on Task Square $D_0$ During Training on Different Data Mixtures.} We visualize the task success rate of a visual diffusion policy averaged across 50 different environment initial conditions on different mixtures of teleoperated demonstrations (TD) and simulated human demonstrations (HD) during training.}
  \label{fig:proof-of-concept}
\end{figure}

To simulate the embodiment gap, we use the KUKA iiwa manipulator to simulate human demonstrations (i.e., demonstrations carried out by a different embodiment) by relabeling the dataset with our pose estimation pipeline, and the Franka Panda for robot demonstrations and the embodiment at test time. As previously mentioned, we use the same colorized gripper, i.e., the Panda gripper on both embodiments. We train diffusion policies implemented by \citet{chi2024diffusionpolicy} out-of-the-box (U-Net with visual observations) for 3000 epochs, which we empirically observe to be sufficient to allow for convergence, following the input arrangement of the diffusion policy out-of-the-box, which include one external view (agentview, with dimension $84\times 84$), one wrist camera view ($84\times 84$), and the end-effector pose, and report the results of training on different mixtures of precisely labeled demonstrations with an embodiment gap and demonstrations with original labels without an embodiment gap. The results are summarized in Table \ref{tab:robosuite_results}, which we discuss in the next paragraph, while Figure \ref{fig:proof-of-concept} visualizes the training curves for different data mixtures on the Square $D_0$ task. In Table \ref{tab:robosuite_results} we choose to display the average of the best consecutive 10 checkpoints instead of the last 10 because the models reach convergence and overfit at different rates when trained on different data mixtures.

Quantitatively, results in Table \ref{tab:robosuite_results} indicate that the success rate of the mixture $0\text{TD}+200\text{HD}$ on average reach $88.1\%$ of the success rate of the mixture $200\text{TD}+0\text{HD}$ in terms of maximum test score, and $87.7\%$ in terms of the average of $10$ best checkpoints, excluding the extreme outlier task Threading. We hypothesize that the results observed for Threading are due to the fact that the point of insertion, which is vital for the success of Threading, lies outside the view of the wrist camera when the insertion is initiated. The wrist camera is embodiment invariant, but the policy has to rely on the agentview camera for Threading. On the other hand, with $50$ robot/same-embodiment demonstrations added, the policy's performance on Threading does scale as we add more human/cross-embodiment demonstrations. This indicates 1) the presence of embodiment invariant views is important for zero-shot cross-embodiment generalization, and therefore the amelioration of occlusion in these views is vital for wide-range task applicability (as explored by \citet{chi2024universal}) and 2) when same-embodiment demonstrations, even a small number, are provided, the addition of precisely labeled cross-embodiment demonstrations increases policy performance despite occlusion in the embodiment-invariant view. All other tasks exhibit the same increased success rate as we increase the number of added cross-embodiment demonstrations on top of $50$ same-embodiment demonstrations. On average across all tasks, policy performance of the best $50$ TD data mixture on average reach $102.6\%$ of the success rate of the mixture $200\text{TD}+0\text{HD}$ in terms of maximum test success rate, and $104.1\%$ in terms of the average of $10$ best checkpoints. Moreover, from the starting epoch of the best consecutive 10 checkpoints we do not observe a definitive trend of slower convergence as the size of the datasets increases, at least for the dataset scale we have tested.

\section{Conclusions, Limitations, Future Work}
In this work we propose a simple and reliable method for extending policy learning data with precisely labeled human demonstrations, and demonstrate its effectiveness in simulated environments. Our results are limited on two fronts: first, the lack of real-world experiments, and second, the lack of an ablation study on larger scale datasets containing more demonstrations or categories of visuomotor data used for training frontier VLA models. The first limitation, caused by our lack of access to equipment, is admittedly nontrivial since our method relies to an extent on precise point cloud reconstruction, which may be subject to harsher hardware limitations in the real world. The second limitation is more appropriately addressed in a larger scale project seeking to propose data strategies for frontier VLA models. We leave both for future work to ameliorate. Otherwise, a topic of future work we suggest is the automated generation of precisely labeled human demonstrations with generative models.

\bibliographystyle{unsrtnat}
\bibliography{references}

\end{document}